%% file: acl2019.tex
\documentclass[11pt,a4paper]{article}
\PassOptionsToPackage{breaklinks}{hyperref}
\usepackage[hyperref]{emnlp-ijcnlp-2019}
\usepackage{times}
\usepackage{latexsym}
\usepackage{amsmath}
\usepackage{booktabs}
\usepackage{amssymb}
\usepackage{url}
\usepackage{multirow}
\usepackage{url}
\usepackage{arydshln}
\aclfinalcopy 

\title{BERT for Coreference Resolution: Baselines and Analysis}

\author{Mandar Joshi $^{\dagger}$ \quad Omer Levy$^{\mathsection}$ \quad Daniel S. Weld$^{\dagger\epsilon}$ \quad Luke Zettlemoyer$^{\dagger\mathsection}$ \\[4pt]
$^{\dagger}$ Allen School of Computer Science \& Engineering, University of Washington, Seattle, WA \\
{\tt \{mandar90,weld,lsz\}@cs.washington.edu}\\[4pt]
$^{\epsilon}$Allen Institute of Artificial Intelligence, Seattle\\
{\tt \{danw\}@allenai.org} \\[4pt]
$^{\mathsection}$ Facebook AI Research, Seattle\\
{\tt \{omerlevy,lsz\}@fb.com} 
}

\date{}
\begin{document}
\maketitle
\input{abstract}

\input{intro}
\input{method}
\input{exp}

\input{discussion}
\input{related}

\bibliography{coref}
\bibliographystyle{acl_natbib}
\end{document}

%% file: abstract.tex
\begin{abstract}
We apply BERT to coreference resolution, achieving strong improvements on the OntoNotes (+3.9 F1) and GAP (+11.5 F1) benchmarks. A qualitative analysis of model predictions indicates that, compared to ELMo and BERT-base, BERT-large is particularly better at distinguishing between related but distinct entities (e.g., President and CEO). However, there is still room for improvement in modeling document-level context, conversations, and mention paraphrasing. Our code and models are publicly available\footnote{\url{https://github.com/mandarjoshi90/coref}}.

\end{abstract}

%% file: intro.tex
\section{Introduction}
Recent BERT-based models have reported dramatic gains on multiple semantic benchmarks including question-answering, natural language inference, and named entity recognition ~\cite{devlin2018bert}. Apart from better bidirectional reasoning, one of BERT's major improvements over previous methods~\cite{Peters2018Elmo,mccann2017learned} is passage-level training,\footnote{Each BERT training example consists of around 512 word pieces, while ELMo is trained on single sentences.}   which allows it to better model longer sequences. 

We fine-tune BERT to coreference resolution, achieving strong improvements on the GAP \cite{webster2018gap} and OntoNotes \cite{Pradhan2012Ontonotes} benchmarks. We present two ways of extending the c2f-coref  model in ~\citet{lee2018higher}. The \emph{independent} variant uses non-overlapping segments each of which acts as an independent instance for BERT. The \emph{overlap} variant splits the document into overlapping segments so as to provide the model with context beyond 512 tokens.
BERT-large improves over ELMo-based c2f-coref 3.9\% on OntoNotes and 11.5\% on GAP (both absolute).

A qualitative analysis of BERT and ELMo-based models (Table ~\ref{tab:errors}) suggests that BERT-large (unlike BERT-base) is remarkably better at distinguishing between related yet distinct entities or concepts (e.g., Repulse Bay and Victoria Harbor). However, both models often struggle to resolve coreferences for cases that require world knowledge (e.g., \emph{the developing story} and \emph{the scandal}). Likewise, modeling pronouns remains difficult, especially in conversations.

We also find that BERT-large benefits from using longer context windows (384 word pieces) while BERT-base performs better with shorter contexts (128 word pieces). Yet, both variants perform much worse with longer context windows (512 tokens) in spite of being trained on 512-size contexts. Moreover, the \emph{overlap} variant, which artificially extends the context window beyond 512 tokens provides no improvement. This indicates that using larger context windows for pretraining might not translate into effective long-range features for a downstream task. Larger models also exacerbate the memory-intensive nature of span representations ~\cite{lee2017end}, which have driven recent improvements in coreference resolution. Together, these observations suggest that there is still room for improvement in modeling document-level context, conversations,
and mention paraphrasing.


%% file: method.tex
\section{Method}
For our experiments, we use the higher-order coreference model in ~\citet{lee2018higher} which is the current state of the art for the English OntoNotes dataset \cite{Pradhan2012Ontonotes}. We refer to this as \emph{c2f-coref} in the paper. 

\subsection{Overview of c2f-coref}
For each mention span $x$, the model learns a distribution $P(\cdot)$ over possible antecedent spans $Y$:
\begin{equation}
    P(y) = \frac{e^{s(x, y)}}{\sum_{y' \in Y} e^{s(x, y')}}
\end{equation}
The scoring function $s(x, y)$ between spans $x$ and $y$ uses fixed-length span representations, $\mathbf{g_x}$ and $\mathbf{g_y}$ to represent its inputs. These consist of a concatenation of three vectors: the two LSTM states of the span endpoints and an attention vector computed over the span tokens.
It computes the score $s(x, y)$ by the mention score of $x$ (i.e. how likely is the span $x$ to be a mention), the mention score of $y$, and the joint compatibility score of $x$ and $y$ (i.e. assuming they are both mentions, how likely are $x$ and $y$ to refer to the same entity). The components are computed as follows:
\begin{align}
    s(x, y) &= s_m(x) + s_m(y) + s_c(x, y) \\
    s_m(x) &= \text{FFNN}_m(\mathbf{g_x}) \\
    s_c(x, y) &= \text{FFNN}_c(\mathbf{g_x}, \mathbf{g_y}, \phi(x, y))
\end{align}
where FFNN$(\cdot)$ represents a feedforward neural network and $\phi(x, y)$ represents speaker and metadata features. These span representations are later refined using antecedent distribution from a span-ranking architecture
as an attention mechanism.


\subsection{Applying BERT}
We replace the entire LSTM-based encoder (with ELMo and GloVe embeddings as input) in c2f-coref with the BERT transformer. We treat the first and last word-pieces (concatenated with the attended version of all word pieces in the span) as span representations.
Documents are split into segments of \texttt{max\_segment\_len}, which we treat as a hyperparameter.  We experiment with two variants of splitting:

\paragraph{Independent}
The \emph{independent} variant uses non-overlapping segments each of which acts as an independent instance for BERT. The representation for each token is limited to the set of words that lie in its segment.  As BERT is trained on sequences of at most 512 word pieces, this variant has limited encoding capacity especially for tokens that lie at the start or end of their segments.

\paragraph{Overlap}
The \emph{overlap} variant splits the document into overlapping segments by creating a $T$-sized segment after every $T/2$ tokens. These segments are then passed on to the BERT encoder independently, and the final token representation is derived by element-wise interpolation of representations from both overlapping segments. 

Let $\mathbf{r_1} \in \mathbb{R}^d $ and $\mathbf{r_2}  \in \mathbb{R}^d $ be the token representations from the overlapping BERT segments. The final representation $\mathbf{r} \in R^d $ is given by: 
\begin{align}
    \mathbf{f} &= \sigma(\mathbf{w}^T [\mathbf{r_1}; \mathbf{r_2}]) \\
    \mathbf{r} &= \mathbf{f} \cdot \mathbf{r_1} + (\mathbf{1} - \mathbf{f}) \cdot \mathbf{r_2}
\end{align}
where $\mathbf{w}  \in \mathbb{R}^{2d \times d}  $ is a trained parameter and $[;]$ represents concatenation. This variant allows the model to artificially increase the context window beyond the \texttt{max\_segment\_len} hyperparameter.

All layers in both model variants are then fine-tuned following \citet{devlin2018bert}. 

%% file: exp.tex
\section{Experiments}
\begin{table*}[t]
\footnotesize
\centering
\setlength{\tabcolsep}{4pt}
\begin{tabular}{lccc@{\hspace{0.5cm}}ccc@{\hspace{0.5cm}}ccc@{\hspace{0.5cm}}c}
\toprule
 & \multicolumn{3}{c}{MUC}& \multicolumn{3}{c}{$\text{B}^3$}& \multicolumn{3}{c}{$\text{CEAF}_{\phi_4}$} \\
 & P & R & F1 & P & R & F1 & P & R & F1 & Avg. F1 \\
 \midrule
\midrule
\citet{Martschat2015Latent}  & 76.7 & 68.1 & 72.2 & 66.1 & 54.2 & 59.6 & 59.5 & 52.3 & 55.7 & 62.5 \\
~\cite{Clark2015Entity}  & 76.1  &69.4 & 72.6  &65.6 & 56.0 & 60.4 & 59.4 & 53.0 & 56.0 & 63.0 \\
\cite{Wiseman2015Learning} & 76.2 & 69.3  & 72.6 & 66.2 & 55.8 & 60.5 & 59.4 & 54.9 & 57.1 & 63.4 \\
\citet{Wiseman2016Global} & 77.5 & 69.8 & 73.4 & 66.8 & 57.0 & 61.5 & 62.1 & 53.9 & 57.7 & 64.2 \\
\citet{clark2016corefRL} &  79.2 & 70.4 & 74.6 & 69.9 & 58.0 & 63.4 & 63.5 & 55.5 & 59.2 & 65.7 \\
e2e-coref \cite{lee2017end}  & 78.4 & 73.4 & 75.8 & 68.6 & 61.8 & 65.0 & 62.7& 59.0 &60.8 &67.2\\
c2f-coref \cite{lee2018higher}  & 81.4 & 79.5 & 80.4 & 72.2 & 69.5 & 70.8 & 68.2 & 67.1 & 67.6 & 73.0 \\
\citet{fei2019end} & {\bf 85.4} & {77.9} & 81.4 & {\bf 77.9} & 66.4 & 71.7 &  70.6 & 66.3 & 68.4 & 73.8 \\
EE \cite{kantor2019coreference} & 82.6 & {\bf 84.1} & 83.4 & 73.3 & {\bf 76.2} & 74.7 & 72.4 & 71.1 & 71.8 & 76.6 \\
\midrule
BERT-base + c2f-coref (independent) & 80.2 & {82.4} & 81.3 & 69.6 & 73.8 & 71.6 & 69.0 & 68.6 & 68.8 & 73.9\\
BERT-base + c2f-coref (overlap) & 80.4 & 82.3 & 81.4 & 69.6 & 73.8 & 71.7 & 69.0 & 68.5 & 68.8 & 73.9 \\
BERT-large + c2f-coref (independent) & 84.7 & {82.4} & {\bf 83.5} & 76.5 & { 74.0} & {\bf 75.3} & {\bf 74.1} & {\bf 69.8} & {\bf 71.9} & \textbf{76.9}\\
BERT-large + c2f-coref (overlap) & { 85.1} & 80.5 & 82.8 & { 77.5} & 70.9 & 74.1 & 73.8 & 69.3 & 71.5 & 76.1\\
\bottomrule
\end{tabular}
\caption{OntoNotes:  BERT improves the c2f-coref model on English by 0.9\% and 3.9\% respectively for base and large variants.
The main evaluation is the average F1 of three metrics -- MUC, $\text{B}^3$, and $\text{CEAF}_{\phi_4}$ on the test set.}
\label{tab:ontonotes}
\end{table*}

We evaluate our BERT-based models on two benchmarks: the paragraph-level GAP dataset~\cite{webster2018gap}, and the document-level English OntoNotes 5.0 dataset~\cite{Pradhan2012Ontonotes}. 
OntoNotes examples are considerably longer and typically require multiple segments to read the entire document.

\paragraph{Implementation and Hyperparameters}
We extend the original Tensorflow implementations of c2f-coref\footnote{\url{http://github.com/kentonl/e2e-coref/}}  and BERT.\footnote{\url{https://github.com/google-research/bert}} We fine tune all models on the OntoNotes English data for 20 epochs using a dropout of 0.3, and learning rates of $1 \times 10^{-5}$ and $2 \times 10^{-4}$ with linear  decay for the BERT parameters and the task parameters respectively. We found that this made a sizable impact of 2-3\% over using the same learning rate for all parameters. 

We  trained separate models with \texttt{max\_segment\_len} of 128, 256, 384, and 512; the models trained on 128 and 384 word pieces performed the best for BERT-base and BERT-large respectively.
As span representations are memory intensive, we truncate documents randomly to eleven segments for BERT-base and 3 for BERT-large during training. Likewise, we use a batch size of 1 document following ~\cite{lee2018higher}. While training the large model requires 32GB GPUs, all models can be tested on 16GB GPUs. We use the cased English variants in all our experiments.

\paragraph{Baselines}
We compare the c2f-coref + BERT system with two main baselines: (1) the original ELMo-based c2f-coref system \cite{lee2018higher}, and (2) its predecessor, e2e-coref \cite{lee2017end}, which does not use contextualized representations. 
In addition to being more computationally efficient than e2e-coref, c2f-coref iteratively refines span representations using attention for higher-order reasoning.

\begin{table}[t]
\footnotesize
\centering
\setlength{\tabcolsep}{4pt}
\begin{tabular}{lcccc}
\toprule
Model & M & F & B & O \\
\midrule
e2e-coref & 67.7 & 60.0 & 0.89 & 64.0\\

c2f-coref & 75.8 & 71.1 & 0.94 & 73.5 \\
BERT + RR ~\citet{LiuRef2019} & 80.3 & 77.4 & \textbf{0.96} & 78.8 \\
BERT-base + c2f-coref   & 84.4 & 81.2 & \textbf{0.96} & 82.8 \\
BERT-large + c2f-coref   & \textbf{86.9} & \textbf{83.0} & 0.95 & \textbf{85.0} \\
\bottomrule
\end{tabular}
\caption{GAP:  BERT improves the c2f-coref model by 11.5\%. The metrics are F1 score on \textbf{M}asculine and \textbf{F}eminine examples, \textbf{O}verall, and a \textbf{B}ias factor (F / M).}
\label{tab:gap_res}
\end{table}

\subsection{Paragraph Level: GAP}
GAP \cite{webster2018gap} is a human-labeled corpus of ambiguous pronoun-name pairs derived from Wikipedia snippets.
Examples in the GAP dataset fit within a single BERT segment, thus eliminating the need for cross-segment inference.
Following ~\citet{webster2018gap}, we trained our BERT-based c2f-coref model on OntoNotes.\footnote{This is motivated by the fact that GAP, with only 4,000 name-pronoun pairs in its dev set, is not intended for full-scale training.}
The predicted clusters were scored against GAP examples according to the official evaluation script. Table~\ref{tab:gap_res} shows that BERT improves c2f-coref by  9\% and 11.5\% for the base and large models respectively. 
These results are in line with large gains reported for a variety of semantic tasks by BERT-based models \cite{devlin2018bert}.


\subsection{Document Level: OntoNotes}
 OntoNotes (English) is  a document-level dataset from the CoNLL-2012 shared task on coreference resolution. It consists of about one million words of newswire, magazine articles, broadcast news, broadcast conversations, web data and conversational
speech data, and the New Testament. The main evaluation is the average F1 of three metrics -- MUC, $\text{B}^3$, and $\text{CEAF}_{\phi_4}$ on the test set according to the official CoNLL-2012 evaluation scripts.

Table~\ref{tab:ontonotes} shows that BERT-base offers an improvement of 0.9\% over the ELMo-based c2f-coref model.   
Given how gains on coreference resolution have been hard to come by as evidenced by the table, this is still a considerable improvement. However, the magnitude of gains is relatively modest considering BERT's arguably better architecture and many more trainable parameters. This is in sharp contrast to how even the base variant of BERT has very substantially improved the state of the art in other tasks. BERT-large, however, improves c2f-coref by the much larger  margin of 3.9\%.
We also observe that the \emph{overlap} variant offers no improvement over \emph{independent}.

Concurrent with our work, ~\citet{kantor2019coreference}, who use higher-order entity-level representations over ``frozen'' BERT features, also report large gains over c2f-coref. While their feature-based approach is more memory efficient, the fine-tuned model seems to yield better results. Also concurrent, SpanBERT ~\cite{joshi2019spanbert}, another self-supervised method, pretrains span representations achieving state of the art results (Avg. F1 79.6) with the \emph{independent} variant.

%% file: discussion.tex
\begin{table*}[t]
\footnotesize
\centering
\begin{tabular}
{llcc}
\toprule
Category & Snippet & \#base & \#large \\
\midrule
\multirow{2}{1.5cm}{Related Entities} & Watch spectacular performances by dolphins and sea lions at the \emph{Ocean Theater}... & 12 & 7\\
& It seems the North Pole and the \underline{Marine Life Center} will also be renovated. & & \\
 \midrule
 Lexical & Over the past 28 years , \emph{the Ocean Park} has basically.. \textbf{The entire park} has been ... & 15 & 9\\
\midrule
Pronouns &  In the meantime , our children need \emph{an education}. \textbf{That}'s all we're asking. & 17 & 13 \\
\midrule
\multirow{2}{1.5cm}{Mention Paraphrasing}& And in case you missed it \emph{the Royals} are here. & 14 & 12\\
& Today Britain's \textbf{Prince Charles and his wife Camilla}... & & \\
\midrule
\multirow{2}{1.5cm}{Conversation} & (Priscilla:) \textbf{My} mother was Thelma Wahl . She was ninety years old ...  &  18 & 16\\
& (Keith:) \emph{Priscilla Scott} is mourning . \emph{Her} mother Thelma Wahl was a resident .. & & \\
\midrule
Misc. & He is \textbf{my}, She is \textbf{my} Goddess , ah  & 17 & 17 \\
\midrule
\emph{Total} & & 93 & 74\\
\bottomrule
\end{tabular}
\caption{Qualitative Analysis: \#base and \#large refers to the number of cluster-level errors on a subset of the OntoNotes English development set. \underline{Underlined} and \textbf{bold-faced} mentions respectively indicate incorrect and missing assignments to \emph{italicized} mentions/clusters. 
The miscellaneous category refers to other errors including (reasonable) predictions that are either missing from the gold data or violate annotation guidelines.
} 
\label{tab:errors}
\end{table*}

\section{Analysis}
We performed a qualitative comparison of ELMo and BERT models (Table \ref{tab:errors}) on the OntoNotes English development set by manually assigning error categories (e.g.,  pronouns, mention paraphrasing) to incorrect predicted clusters.\footnote{Each incorrect cluster can belong to multiple categories.} Overall, we found 93 errors for BERT-base and 74 for BERT-large from the same 15 documents.

\begin{table}[t]
\footnotesize
\centering
\setlength{\tabcolsep}{4pt}
\begin{tabular}{rcrcc}
\toprule
Doc length & \#Docs & Spread & F1 (base) & F1 (large)\\
\midrule
0 - 128 & 48 & 37.3 & 80.6 & 84.5 \\
128 - 256 & 54 & 71.7 & 80.0 & 83.0\\
256 - 512 & 74 & 109.9 & 78.2 & 80.0\\
512 - 768 & 64 & 155.3 & 76.8 & 80.2\\
768 - 1152 &  61 & 197.6 & 71.1 & 76.2\\
1152+ & 42 & 255.9& 69.9 & 72.8\\
\midrule
All & 343  & 179.1 & 74.3 & 77.3\\
\bottomrule
\end{tabular}
\caption{Performance on the English OntoNotes dev set generally drops as the document length increases. Spread is measured as the average number of tokens between the first and last mentions in a cluster.}
\label{tab:num_seg_res}
\end{table}

\begin{table}[t]
\footnotesize
\centering
\setlength{\tabcolsep}{4pt}
\begin{tabular}{lcc}
\toprule
Segment Length & F1 (BERT-base) & F1 (BERT-large)\\
\midrule
128 & 74.4 & 76.6 \\
256 & 73.9 & 76.9\\
384 & 73.4 & 77.3\\
450 &  72.2 & 75.3\\
512 &  70.7 & 73.6\\
\bottomrule
\end{tabular}
\caption{Performance on the English OntoNotes dev set with varying values for \texttt{max\_segment\_len}. Neither model is able  to effectively exploit larger segments; they perform especially badly  when  \texttt{maximum\_segment\_len} of 512 is used.}
\label{tab:seg_len_res}
\end{table}

\paragraph{Strengths}
 We did not find salient qualitative differences between ELMo and BERT-base models, which is consistent with the quantitative results (Table \ref{tab:ontonotes}). BERT-large improves over BERT-base in a variety of ways including pronoun resolution and lexical matching (e.g., \emph{race track} and \emph{track}). In particular,  the BERT-large variant is better at distinguishing related, but distinct, entities.  Table \ref{tab:errors} shows several examples where the BERT-base variant merges distinct entities (like \emph{Ocean Theater} and \emph{Marine Life Center}) into a single cluster.  BERT-large seems to be able to avoid such merging on a more regular basis.

\paragraph{Weaknesses}
An analysis of errors on the OntoNotes English development set suggests that better modeling of document-level context, conversations, and entity paraphrasing might further improve the state of the art.

\emph{Longer documents} in OntoNotes generally contain larger and more spread-out clusters. We focus on three observations -- (a) Table \ref{tab:num_seg_res} shows how models perform distinctly worse on longer documents, (b) both models are unable to use larger segments more effectively (Table \ref{tab:seg_len_res}) and perform worse when the \texttt{max\_segment\_len} of 450 and 512 are used, and, (c) using overlapping segments to provide additional context does not improve results (Table \ref{tab:ontonotes}). Recent work ~\cite{joshi2019spanbert} suggests that BERT's inability to use longer sequences effectively is likely a by-product pretraining on short sequences for a vast majority of updates.

Comparing preferred segment lengths for base and large variants of BERT indicates that larger models might better encode longer contexts. However, larger models also exacerbate the memory-intensive nature of span representations,\footnote{We required a 32GB GPU to finetune BERT-large.} which have driven recent improvements in coreference resolution. These observations suggest that future research in pretraining methods should look at more \emph{effectively} encoding document-level context using sparse representations ~\cite{child2019generating}. 

Modeling \emph{pronouns} especially in the context of conversations (Table \ref{tab:errors}), continues to be difficult for all models, perhaps partly because c2f-coref does very little to model dialog structure of the document.
Lastly, a considerable number of errors suggest that models are still unable to resolve cases requiring \emph{mention paraphrasing}. For example, bridging \emph{the Royals} with \emph{Prince Charles and his wife Camilla} likely requires pretraining models to encode relations between entities, especially considering that such learning signal is rather sparse in the training set.

%% file: related.tex
\section{Related Work}
Scoring span or mention pairs has perhaps been one of the most dominant paradigms in coreference resolution. The base coreference model used in this paper from ~\citet{lee2018higher} belongs to this family of models ~\cite{Ng2002Identifying,Bengtson2008Understanding,denis2008specialized,fernandes2012latest,durrett2013easy,Wiseman2015Learning,clark2016corefRL,lee2017end}. 

More recently, advances in coreference resolution and other NLP tasks have been driven by unsupervised contextualized representations ~\cite{Peters2018Elmo,devlin2018bert,mccann2017learned,joshi2019spanbert,liu2019roberta}. Of these, BERT ~\cite{devlin2018bert} notably uses pretraining on passage-level sequences (in conjunction with a bidirectional masked language modeling objective) to more effectively model long-range dependencies. SpanBERT ~\cite{joshi2019spanbert} focuses on pretraining span representations achieving current state of the art results on  OntoNotes with the \emph{independent} variant.

%% file: acl2019.bbl
\begin{thebibliography}{23}
\expandafter\ifx\csname natexlab\endcsname\relax\def\natexlab#1{#1}\fi

\bibitem[{Bengtson and Roth(2008)}]{Bengtson2008Understanding}
Eric Bengtson and Dan Roth. 2008.
\newblock \href {http://dl.acm.org/citation.cfm?id=1613715.1613756}
  {Understanding the value of features for coreference resolution}.
\newblock In \emph{Proceedings of the Conference on Empirical Methods in
  Natural Language Processing}, EMNLP '08, pages 294--303, Stroudsburg, PA,
  USA. Association for Computational Linguistics.

\bibitem[{Child et~al.(2019)Child, Gray, Radford, and
  Sutskever}]{child2019generating}
Rewon Child, Scott Gray, Alec Radford, and Ilya Sutskever. 2019.
\newblock \href {https://arxiv.org/abs/1904.10509} {Generating long sequences
  with sparse transformers}.
\newblock \emph{arXiv preprint arXiv:1904.10509}.

\bibitem[{Clark and Manning(2015)}]{Clark2015Entity}
Kevin Clark and Christopher~D. Manning. 2015.
\newblock \href {https://doi.org/10.3115/v1/P15-1136} {Entity-centric
  coreference resolution with model stacking}.
\newblock In \emph{Proceedings of the 53rd Annual Meeting of the Association
  for Computational Linguistics and the 7th International Joint Conference on
  Natural Language Processing (Volume 1: Long Papers)}, pages 1405--1415.
  Association for Computational Linguistics.

\bibitem[{Clark and Manning(2016)}]{clark2016corefRL}
Kevin Clark and Christopher~D. Manning. 2016.
\newblock \href {https://doi.org/10.18653/v1/D16-1245} {Deep reinforcement
  learning for mention-ranking coreference models}.
\newblock In \emph{Proceedings of the 2016 Conference on Empirical Methods in
  Natural Language Processing}, pages 2256--2262. Association for Computational
  Linguistics.

\bibitem[{Denis and Baldridge(2008)}]{denis2008specialized}
Pascal Denis and Jason Baldridge. 2008.
\newblock \href {http://aclweb.org/anthology/D08-1069} {Specialized models and
  ranking for coreference resolution}.
\newblock In \emph{Proceedings of the 2008 Conference on Empirical Methods in
  Natural Language Processing}, pages 660--669. Association for Computational
  Linguistics.

\bibitem[{Devlin et~al.(2019)Devlin, Chang, Lee, and
  Toutanova}]{devlin2018bert}
Jacob Devlin, Ming-Wei Chang, Kenton Lee, and Kristina Toutanova. 2019.
\newblock \href {https://doi.org/10.18653/v1/N19-1423} {{BERT}: Pre-training of
  deep bidirectional transformers for language understanding}.
\newblock In \emph{Proceedings of the 2019 Conference of the North {A}merican
  Chapter of the Association for Computational Linguistics: Human Language
  Technologies, Volume 1 (Long and Short Papers)}, pages 4171--4186,
  Minneapolis, Minnesota. Association for Computational Linguistics.

\bibitem[{Durrett and Klein(2013)}]{durrett2013easy}
Greg Durrett and Dan Klein. 2013.
\newblock \href {http://aclweb.org/anthology/D13-1203} {Easy victories and
  uphill battles in coreference resolution}.
\newblock In \emph{Proceedings of the 2013 Conference on Empirical Methods in
  Natural Language Processing}, pages 1971--1982. Association for Computational
  Linguistics.

\bibitem[{Fei et~al.(2019)Fei, Li, Li, and Li}]{fei2019end}
Hongliang Fei, Xu~Li, Dingcheng Li, and Ping Li. 2019.
\newblock \href {https://www.aclweb.org/anthology/P19-1064} {End-to-end deep
  reinforcement learning based coreference resolution}.
\newblock In \emph{Proceedings of the 57th Annual Meeting of the Association
  for Computational Linguistics}, pages 660--665, Florence, Italy. Association
  for Computational Linguistics.

\bibitem[{Fernandes et~al.(2012)Fernandes, dos Santos, and
  Milidi{\'u}}]{fernandes2012latest}
Eraldo Fernandes, C{\'i}cero dos Santos, and Ruy Milidi{\'u}. 2012.
\newblock \href {http://aclweb.org/anthology/W12-4502} {Latent structure
  perceptron with feature induction for unrestricted coreference resolution}.
\newblock In \emph{Joint Conference on EMNLP and CoNLL - Shared Task}, pages
  41--48. Association for Computational Linguistics.

\bibitem[{Joshi et~al.(2019)Joshi, Chen, Liu, Weld, Zettlemoyer, and
  Levy}]{joshi2019spanbert}
Mandar Joshi, Danqi Chen, Yinhan Liu, Daniel~S. Weld, Luke Zettlemoyer, and
  Omer Levy. 2019.
\newblock \href {https://arxiv.org/abs/1907.10529} {{SpanBERT}: Improving
  pre-training by representing and predicting spans}.
\newblock \emph{arXiv preprint arXiv:1907.10529}.

\bibitem[{Kantor and Globerson(2019)}]{kantor2019coreference}
Ben Kantor and Amir Globerson. 2019.
\newblock \href {https://www.aclweb.org/anthology/P19-1066} {Coreference
  resolution with entity equalization}.
\newblock In \emph{Proceedings of the 57th Annual Meeting of the Association
  for Computational Linguistics}, pages 673--677, Florence, Italy. Association
  for Computational Linguistics.

\bibitem[{Lee et~al.(2017)Lee, He, Lewis, and Zettlemoyer}]{lee2017end}
Kenton Lee, Luheng He, Mike Lewis, and Luke Zettlemoyer. 2017.
\newblock \href {https://doi.org/10.18653/v1/D17-1018} {End-to-end neural
  coreference resolution}.
\newblock In \emph{Proceedings of the 2017 Conference on Empirical Methods in
  Natural Language Processing}, pages 188--197, Copenhagen, Denmark.
  Association for Computational Linguistics.

\bibitem[{Lee et~al.(2018)Lee, He, and Zettlemoyer}]{lee2018higher}
Kenton Lee, Luheng He, and Luke Zettlemoyer. 2018.
\newblock \href {https://doi.org/10.18653/v1/N18-2108} {Higher-order
  coreference resolution with coarse-to-fine inference}.
\newblock In \emph{Proceedings of the 2018 Conference of the North {A}merican
  Chapter of the Association for Computational Linguistics: Human Language
  Technologies, Volume 2 (Short Papers)}, pages 687--692, New Orleans,
  Louisiana. Association for Computational Linguistics.

\bibitem[{Liu et~al.(2019{\natexlab{a}})Liu, Zettlemoyer, and
  Eisenstein}]{LiuRef2019}
Fei Liu, Luke Zettlemoyer, and Jacob Eisenstein. 2019{\natexlab{a}}.
\newblock \href {https://arxiv.org/abs/1902.01541} {The referential reader: A
  recurrent entity network for anaphora resolution}.
\newblock In \emph{Proceedings of the 57th Annual Meeting of the Association
  for Computational Linguistics}, pages 5918--5925, Florence, Italy.

\bibitem[{Liu et~al.(2019{\natexlab{b}})Liu, Ott, Goyal, Du, Joshi, Chen, Levy,
  Lewis, Zettlemoyer, and Stoyanov}]{liu2019roberta}
Yinhan Liu, Myle Ott, Naman Goyal, Jingfei Du, Mandar Joshi, Danqi Chen, Omer
  Levy, Mike Lewis, Luke Zettlemoyer, and Veselin Stoyanov. 2019{\natexlab{b}}.
\newblock \href {https://arxiv.org/abs/1907.11692} {Ro{BERT}a: A robustly
  optimized {BERT} pretraining approach}.

\bibitem[{Martschat and Strube(2015)}]{Martschat2015Latent}
Sebastian Martschat and Michael Strube. 2015.
\newblock \href {https://transacl.org/ojs/index.php/tacl/article/view/604}
  {Latent structures for coreference resolution}.
\newblock \emph{Transactions of the Association for Computational Linguistics},
  3:405--418.

\bibitem[{McCann et~al.(2017)McCann, Bradbury, Xiong, and
  Socher}]{mccann2017learned}
Bryan McCann, James Bradbury, Caiming Xiong, and Richard Socher. 2017.
\newblock \href {https://arxiv.org/abs/1708.00107} {Learned in translation:
  Contextualized word vectors}.
\newblock In \emph{Advances in Neural Information Processing Systems}, pages
  6297--6308.

\bibitem[{Ng and Cardie(2002)}]{Ng2002Identifying}
Vincent Ng and Claire Cardie. 2002.
\newblock \href {https://doi.org/10.3115/1072228.1072367} {Identifying
  anaphoric and non-anaphoric noun phrases to improve coreference resolution}.
\newblock In \emph{Proceedings of the 19th International Conference on
  Computational Linguistics - Volume 1}, COLING '02, pages 1--7, Stroudsburg,
  PA, USA. Association for Computational Linguistics.

\bibitem[{Peters et~al.(2018)Peters, Neumann, Iyyer, Gardner, Clark, Lee, and
  Zettlemoyer}]{Peters2018Elmo}
Matthew Peters, Mark Neumann, Mohit Iyyer, Matt Gardner, Christopher Clark,
  Kenton Lee, and Luke Zettlemoyer. 2018.
\newblock \href {https://doi.org/10.18653/v1/N18-1202} {Deep contextualized
  word representations}.
\newblock In \emph{Proceedings of the 2018 Conference of the North American
  Chapter of the Association for Computational Linguistics: Human Language
  Technologies, Volume 1 (Long Papers)}, pages 2227--2237. Association for
  Computational Linguistics.

\bibitem[{Pradhan et~al.(2012)Pradhan, Moschitti, Xue, Uryupina, and
  Zhang}]{Pradhan2012Ontonotes}
Sameer Pradhan, Alessandro Moschitti, Nianwen Xue, Olga Uryupina, and Yuchen
  Zhang. 2012.
\newblock \href {http://aclweb.org/anthology/W12-4501} {Conll-2012 shared task:
  Modeling multilingual unrestricted coreference in ontonotes}.
\newblock In \emph{Joint Conference on EMNLP and CoNLL - Shared Task}, pages
  1--40. Association for Computational Linguistics.

\bibitem[{Webster et~al.(2018)Webster, Recasens, Axelrod, and
  Baldridge}]{webster2018gap}
Kellie Webster, Marta Recasens, Vera Axelrod, and Jason Baldridge. 2018.
\newblock \href {https://doi.org/10.1162/tacl_a_00240} {Mind the {GAP}: A
  balanced corpus of gendered ambiguous pronouns}.
\newblock \emph{Transactions of the Association for Computational Linguistics},
  6:605--617.

\bibitem[{Wiseman et~al.(2015)Wiseman, Rush, Shieber, and
  Weston}]{Wiseman2015Learning}
Sam Wiseman, Alexander~M. Rush, Stuart Shieber, and Jason Weston. 2015.
\newblock \href {https://doi.org/10.3115/v1/P15-1137} {Learning anaphoricity
  and antecedent ranking features for coreference resolution}.
\newblock In \emph{Proceedings of the 53rd Annual Meeting of the Association
  for Computational Linguistics and the 7th International Joint Conference on
  Natural Language Processing (Volume 1: Long Papers)}, pages 1416--1426.
  Association for Computational Linguistics.

\bibitem[{Wiseman et~al.(2016)Wiseman, Rush, and Shieber}]{Wiseman2016Global}
Sam Wiseman, Alexander~M. Rush, and Stuart~M. Shieber. 2016.
\newblock \href {https://doi.org/10.18653/v1/N16-1114} {Learning global
  features for coreference resolution}.
\newblock In \emph{Proceedings of the 2016 Conference of the North American
  Chapter of the Association for Computational Linguistics: Human Language
  Technologies}, pages 994--1004. Association for Computational Linguistics.

\end{thebibliography}
